\documentclass{article}

\PassOptionsToPackage{numbers, compress}{natbib}

\usepackage[nonatbib, preprint]{neurips_2025}

\usepackage[utf8]{inputenc} 
\usepackage[T1]{fontenc}    
\usepackage{hyperref}       
\usepackage{url}            
\usepackage{booktabs}       
\usepackage{amsfonts}       
\usepackage{nicefrac}       
\usepackage{microtype}      
\usepackage{xcolor}         

\usepackage{graphicx}
\usepackage{makecell}
\usepackage{multirow}

\title{Attention Is Not All You Need: The Importance of Feedforward Networks in Transformer Models}

\author{%
  Isaac Gerber \\
  Johns Hopkins University\\
  \texttt{igerber2@jhu.edu } \\
}

\begin{document}

\maketitle

\begin{abstract}
Decoder-only transformer networks have become incredibly popular for language modeling tasks. State-of-the-art models can have over a hundred transformer blocks, containing billions of trainable parameters, and are trained on trillions of tokens of text. Each transformer block typically consists of a multi-head attention (MHA) mechanism and a two-layer fully connected feedforward network (FFN). In this paper, we examine the importance of the FFN during the model pre-training process through a series of experiments, confirming that the FFN is important to model performance. Furthermore, we show that models using a transformer block configuration with three-layer FFNs with fewer such blocks outperform the standard two-layer configuration delivering lower training loss with fewer total parameters in less time.
\end{abstract}

\section{Introduction}
The decoder-only transformer architecture popularized by OpenAI's GPT models \cite{Radford_Narasimhan_Salimans_Sutskever_2018, Radford_Wu_Child_Luan_Amodei_Sutskever_2019} consists of many stacked transformer blocks that each include two key components: a self-attention mechanism, typically a form of multi-head attention (MHA), followed by a two-layer feedforward network (FFN). Although significant efforts have focused on optimizing the MHA, less attention has been paid to the FFN.

A standard transformer block has more trainable parameters in the FFN than in the MHA, as explored in Section \ref{background}. Given that top large language models (LLMs) have billions of trainable parameters and are pre-trained on trillions of tokens of text \cite{llama_3}, allocating a parameter budget efficiently is vital.

On the other hand, pre-LLM research into multilayer perceptions (MLPs) show that FFNs with a hidden layer can act as universal function approximators \cite{HORNIK1989359}. For this reason, it is possible that adding a third, hidden layer to the transformer block FFNs will improve model performance.

In this paper, we evaluate the importance of the FFN within the transformer block through a series of experiments, examining the performance of models with three linear layers, two linear layers (the baseline), one linear layer, and zero linear layers per block. We further determine if performance is driven by the FFN or the total number of trainable parameters, by changing the model dimension size ($d_{model}$) or the number of transformer blocks to keep the number of trainable parameters roughly constant.

\section{Background}\label{background}
Vaswani et al. \cite{Vaswani_Shazeer_Parmar_Uszkoreit_Jones_Gomez_Kaiser_Polosukhin_2017} introduced the transformer model architecture, consisting of input embeddings, stacks of transformer blocks which contain an encoder and decoder, followed by output probabilities. Current approaches to transformer-based language models typically omit the encoder, stacking decoder-only transformer blocks \cite{wang2022languagemodelarchitecturepretraining}. 

The decoder transformer block has two main components: MHA and an FFN. The MHA mechanism learns three sets of weights, $W^q$, $W^k$, $W^v$, corresponding to the queries, key, and values, further split into $h$ heads. Combined across all such heads, the dimensions of each weight matrix are $d_{model} \times d_{model}$, resulting in $3 \times d_{model} \times d_{model}$ learnable weights. The standard FFN portion of the transformer block has two linear layers: a $4d_{model} \times d_{model}$ layer and a $d_{model} \times 4d_{model}$ layer, resulting in $8 \times d_{model} \times d_{model}$ learnable weights. When compared, we see that the standard transformer block learns parameters in an 8:3 ratio between the FFN and MHA, respectively.

\section{Related Work}
Sukhbaatar et al. \cite{Sukhbaatar_Grave_Lample_Jegou_Joulin_2019} show that complete removal of the linear layers and associated activation function significantly degrades model performance. However, their approach does not consider the inclusion of a single linear layer, nor does it examine model performance while retaining an activation function. He et al. \cite{he2024matterstransformersattentionneeded} perform ablation studies on layers and full transformer blocks. However, their work only considers an already pre-trained model, Llama-2, and does not examine the effect on the pre-training process. 
\cite{sridhar2022trimberttailoringberttradeoffs} and \cite{sridhar2023undividedattentionintermediatelayers} explore removing the intermediate layers between transformer blocks in the BERT model architecture, but do not explore the effect of the FFN within the transformer block itself.

\cite{NEURIPS2024_2d8f2351} studies the theoretical effect of transformer components, validating the results with experiments. They study the effect of the FFN width, finding that wider FFNs can lead to better model performance, but do not consider changing the number of layers in the FFN. \cite{geva-etal-2021-transformer} shows that the FFN in transformer models act as key-value memories, but do not study the effect of changing the number of layers on model performance. \cite{one-wide-ffn} examines replacing the block-specific FFNs in an encoder+decoder transformer model with shared FFNs, finding that FFNs can be shared across either the encoder or decoder or both. However, their experiments on decoder-only models are limited, and they look specifically at the machine translation task, not LLM pre-training.

\section{Methods}\label{methods}
\subsection{Model Architectures and Implementation Details}\label{implementation}
\begin{figure}
    \centering
    \includegraphics[width=1.0\linewidth]{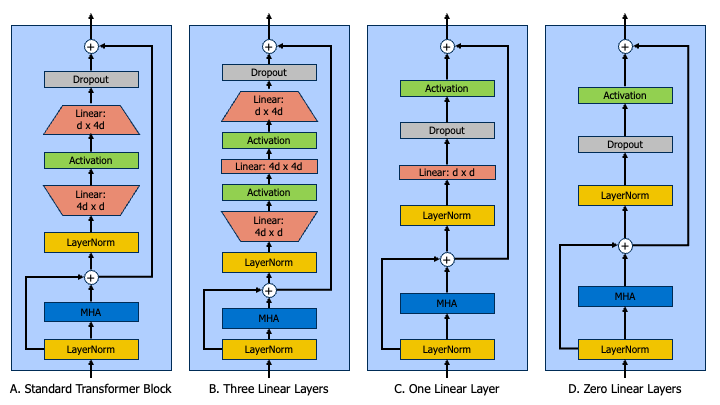}
    \caption{Transformer block architecture. \textbf{A} depicts a standard transformer block with two linear layers, \textbf{B} a transformer block with three linear layers, \textbf{C} a transformer block with a single linear layer, and \textbf{D} a transformer block with zero linear layers. The arrows at the bottom and top of each block indicate input and output from the previous and to the next sections of the models, respectively. In the linear layers, "d" refers to the model dimension size, $d_{model}$. A transformer model stacks many transformer blocks.}
    \label{fig:arch}
\end{figure}
Figure \ref{fig:arch} shows the standard transformer block architecture and the three variants that we tested. In the one- and zero-layer variants, the dropout layer comes before the activation layer, whereas in the standard version, the dropout layer is the last. As dropout zeros out the output from some parameters during training, and activation functions map zero to zero, these are functionally equivalent. 

Our base model configuration contains two linear layers in the FFN, 24 transformer blocks in the overall model, and a $d_{model}$ value of 1024, resulting in 323 million trainable parameters. We chose this base configuration to match the "GPT-3 Medium" architecture \cite{brown2020languagemodelsfewshotlearners}.

We considered variations with three linear layers, one linear layer, and zero linear layers. For each, we experimented with constant depth (number of transformer blocks) and $d_{model}$ values, as well as varying either $d_{model}$ or model depth to keep the parameter count within 3\% of the baseline model.

We used GELU \cite{Hendrycks_Gimpel_2023} as our activation function and a dropout proportion of 0.1. The MHA layers have 16 heads. We used Xavier uniform initialization \cite{pmlr-v9-glorot10a} for the MHA and linear layers, and normal initialization for the embedding layers.

Our model uses the HuggingFace Tokenizer\footnote{https://huggingface.co/docs/transformers/en/main\_classes/tokenizer} with a maximum vocabulary of 10,000. We trained the models on a single epoch across each set of training data. We use the AdamW optimizer with a cosine learning rate decay with 300 warm-up steps with a maximum learning rate of $1.5^{-4}.$ We used a sequence length of 256 and 16 sequences per batch. Models were both trained and evaluated using one NVIDIA A100 in Google Colab\footnote{https://colab.research.google.com/}.

\subsection{Data Sets}\label{data_sets}
We trained and evaluated our models on the "Booksum Complete Cleaned"{\footnote{https://huggingface.co/datasets/ubaada/booksum-complete-cleaned}} a processed version of \cite{kryściński2022booksumcollectiondatasetslongform} (Booksum), available via the BSD-3 License, and "Wikitext 103 v1" \cite{merity2016pointer} (Wikitext) data sets, available via Creative Commons Attribution-ShareAlike License. Both data sets are divided into training, validation, and test sets.

Booksum's corpus contains 144,846 sequences of training data and 24,220 of test data. With a batch size of 16, this results in 9,053 and 1,516 batches during training and evaluation, respectively. Wikitext's corpus contains 510,089 training sequences and 1,212 test sequences, resulting in 31,881 training batches and 76 evaluation batches.

\subsection{Experimental Design}
For each data set we trained a tokenizer on the training data and used that tokenizer on the corresponding training and test data. We then trained each model configuration on each set of tokenized training data and evaluated it on the corresponding test data. Finally, we calculated the mean cross-entropy loss on each sequence, reporting the mean and standard error across all batches. 

\section{Results} \label{results}
\begin{table}[ht]
    \centering
    \begin{tabular}{ |c|c|c|c!{\vrule width 1.5pt}c|c| }
    \hline
    \multicolumn{4}{|c!{\vrule width 1.5pt}}{\textbf{Model Configuration}} & \multicolumn{2}{c|}{\textbf{Loss}} \\

    \makecell{\# Linear Layers\\per Block} & \makecell{\# Transformer\\Blocks} & $d_{model}$ & \# Params & Booksum & Wikitext \\
    \noalign{\hrule height 1.5pt}
    3 & 24 & 1024 & 726M & 4.232 (0.002) & 3.018 (0.014) \\
    3 & 24 & 672 & 318M & 4.279 (0.002) & 3.078 (0.014) \\
    3 & 10 & 1024 & 314M & \textbf{4.208} (0.002) & \textbf{2.987} (0.014) \\
    \hline
    2 (baseline) & 24 & 1024 & 323M & 4.259 (0.002) & 3.001 (0.014) \\
    \hline
    1 & 24 & 1024 & 147M & 4.405 (0.002) & 3.185 (0.014) \\
    1 & 24 & 1568 & 327M & 4.292 (0.002) & 3.077 (0.014) \\
    1 & 57 & 1024 & 320M & 4.401 (0.002) & 3.128 (0.014) \\
    \hline
    0 & 24 & 1024 & 121M & 4.610 (0.002) & 3.446 (0.013) \\
    0 & 24 & 1728 & 322M & 4.495 (0.002) & 3.333 (0.013) \\
    0 & 72 & 1024 & 322M & 4.720 (0.002) & 3.506 (0.013) \\
    \hline
    \end{tabular}
    \caption{Model configuration and loss (standard error) on Booksum and Wikitext test sets.}
    \label{tab:overall_loss}
\end{table}

Table \ref{tab:overall_loss} summarizes model configuration and performance across both data sets. The three-layer FFN with 10 transformer blocks outperforms all other configurations on both data sets. That said, its performance on the Wikitext data set is not statistically different from the baseline configuration at the 5\% significance level. 

The three-layer FFN with 24 blocks and $d_{model}=1024$ also outperforms the baseline on the Booksum data set (statistically significant at 5\%), but underperforms on the Wikitext data set (not statistically significant). Given the over doubling of parameters when adding a layer to the FFN, from 323M to 726M, the larger models may be overfitting the training data.

In general, models with more linear layers outperform the models with fewer linear layers on both data sets. For both the one-layer and zero-layer FFNs, the higher-dimension models outperform their deeper network counterparts, suggesting an essential balance between model dimensionality and depth that could be explored in future experiments.

\section{Discussion}\label{discussion}

\subsection{Limitations}\label{limitations}
We experiment on only two data sets. Although our findings are consistent across both data sets, it is possible that they do not hold when considering other data sets. Due to computation constraints, we only train and evaluate each model architecture once on each data set. Replications may yield different results.

We only consider the standard MHA mechanism. Many current models use updated attention mechanisms such as FlashAttention \cite{dao2022flashattentionfastmemoryefficientexact} or Star Attention \cite{acharya2025starattentionefficientllm}. It is possible that the findings here do not hold when using other attention mechanisms.
 
Deep neural networks have many tunable hyperparameters but are expensive to train, limiting our ability to validate the best hyperparameters. More time spent validating the best hyperparameters for each model configuration could yield different performance metrics than we achieved. For example, given the variability we saw during training, different warm-up values for the learning rate scheduler per architecture variant might deliver better performance. Experimenting with the balance between model depth and dimensionality could be helpful. Additionally, deeper networks, trained on more data, could confirm if the trends we observe hold at scale. 

Finally, though we evaluated the performance based on cross-entropy loss, a more specialized metric based on the final downstream task may give a better signal. For example, performing supervised fine tuning and assessing relative performance using the MMLU metric \cite{hendryckstest2021} would demonstrate if the change in pre-training cross-entropy loss is consistent with overall model quality.

\subsection{Training}\label{training}
In addition to the performance differences noted in Section \ref{results}, models with more linear layers were generally more stable when training, suffering less from vanishing gradients. Our initial attempts used a max learning rate of $2.5^{-4}$, the ReLU activation function, and uniform weight initialization, which worked for the baseline configuration, but led to vanishing gradients in the models with fewer linear layers. Adjusting all three of these to the values shown in Section \ref{implementation} was necessary to train a stable model.

\begin{figure}
    \centering
    \includegraphics[width=1.0\linewidth]{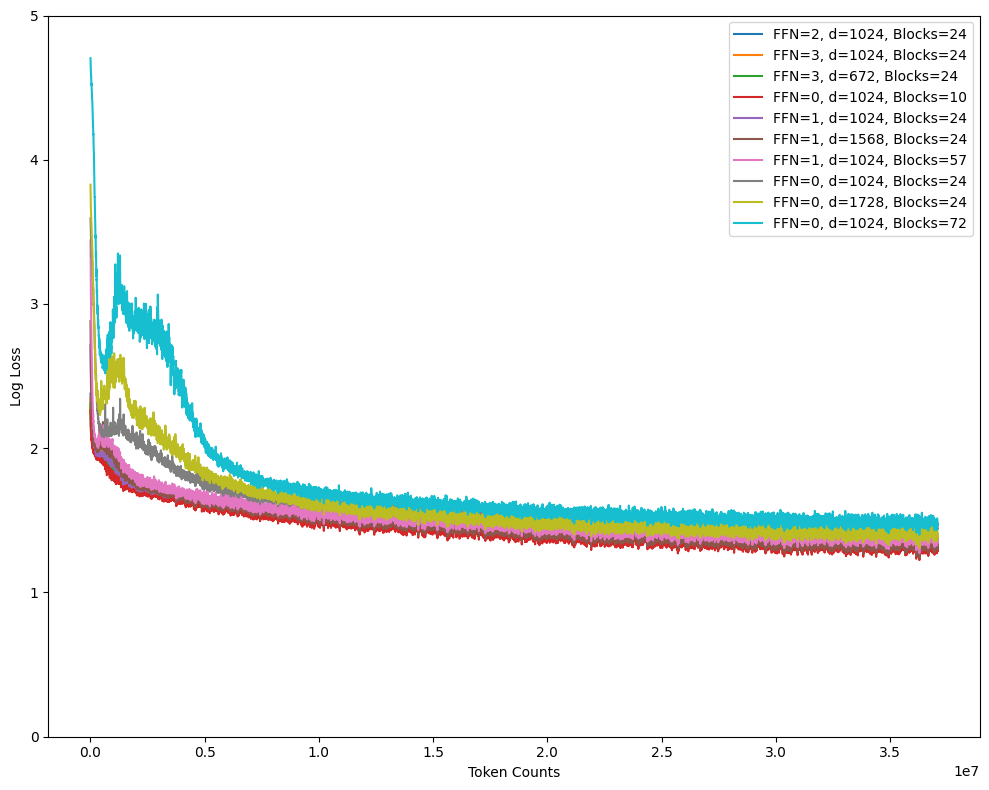}
    \caption{Training loss on the Booksum data set, log scale.}
    \label{fig:training_loss}
\end{figure}

Figure \ref{fig:training_loss} shows the log loss when training on the Booksum data set. The fewer linear layers, the larger the initial training loss. The scale is notable with the two-layer network starting under 2.5, the one-layer networks starting between 2.8 and 3.5, and the zero-layer networks starting between 3.5 and 4.8. Adding dimensions or increasing the network depth increases the initial training loss. Additionally, the fewer the linear layers and the deeper the model/more dimensions in the model, the more variance we see within the first 500k tokens. Further experiments adjusting the number of warm-up steps in the learning rate scheduler for each model variant could result in more stable training and better performance.

\subsection{Computation Time}\label{compute}
\begin{table}[ht]
    \centering
    \begin{tabular}{ |c|c|c|c!{\vrule width 1.5pt}c|c| }
    \hline
    \multicolumn{4}{|c!{\vrule width 1.5pt}}{\textbf{Model Configuration}} & \multicolumn{2}{c|}{\textbf{GPU Hours}} \\

    \makecell{\# Linear Layers\\per Block} & \makecell{\# Transformer\\Blocks} & $d_{model}$ & \# Params & Booksum & Wikitext \\
    \noalign{\hrule height 1.5pt}
    3 & 24 & 1024 & 726M & 2.82 & 9.93 \\
    3 & 24 & 672 & 318M & 1.43 & 5.07 \\
    3 & 10 & 1024 & 314M & 1.23 & 4.35 \\
    \hline
    2 (baseline) & 24 & 1024 & 323M & 1.42 & 4.98 \\
    \hline
    1 & 24 & 1024 & 147M & 0.77 & 2.70 \\
    1 & 24 & 1568 & 327M & 1.45 & 5.13 \\
    1 & 57 & 1024 & 320M & 1.68 & 5.93 \\
    \hline
    0 & 24 & 1024 & 121M & \textbf{0.67} & \textbf{2.37} \\
    0 & 24 & 1728 & 322M & 1.38 & 4.90 \\
    0 & 72 & 1024 & 322M & 1.82 & 6.40 \\
    \hline
    \end{tabular}
    \caption{Training time on an A100 GPU in hours.}
    \label{tab:compute_table}
\end{table}

Our primary experiments reported in Section \ref{results} took 66.43 hours of compute for model training. We detail the training time by configuration and data set in Table \ref{compute}. 

In general, models with more parameters take longer to train, and deeper models take longer to train than shallower models.. We see this reflected in the largest model, the three-layer model with 24 blocks and $d_{model}$ of 1024 taking the longest to train while the smallest model, the zero-layer models with 24 blocks and $d_{model}$ of 1024, takes the shortest time.

The best performing model in Section \ref{results}, the three-layer model with 10 blocks and $d_{model}$ of 1024 takes about 13\% less time to train than the baseline model configuration.

\subsection{Two-Layer FFN Exploration}\label{two-layer-exploration}
\begin{table}[ht]
    \centering
    \begin{tabular}{ |c|c|c|c!{\vrule width 1.5pt}c|c| }
    \hline
    \multicolumn{4}{|c!{\vrule width 1.5pt}}{\textbf{Model Configuration}} & \multicolumn{2}{c|}{\textbf{Loss}} \\

    FFN Size & \makecell{\# Transformer\\Blocks} & $d_{model}$ & \# Params & Booksum & Wikitext \\
    \noalign{\hrule height 1.5pt}
    $4d$ (baseline) & 24 & 1024 & 323M & \textbf{4.259} (0.002) & \textbf{3.001} (0.014) \\
    $4d$ & 24 & 672 & 144M & 4.334 (0.002) & 3.135 (0.013) \\
    $4d$ & 10 & 1024 & 147M & 4.260 (0.002) & 3.062 (0.014) \\
    \hline
    $2d$ & 24 & 1024 & 222M & 4.288 (0.002) & 3.060 (0.014) \\
    $2d$ & 24 & 1248 & 324M & 4.268 (0.002) & 3.013 (0.014) \\
    \hline
    \end{tabular}
    \caption{Model configuration and loss (standard error) on Booksum and Wikitext test sets for two-layer FFNs with different dimension sizes.}
    \label{tab:ffn_2}
\end{table}

Table \ref{tab:ffn_2} explores the performance of two-layer FFNs. The first row is replicated from the first row in Table \ref{tab:overall_loss}, which represents the standard model configuration of a $4d_{model} \times d_{model}$ layer and a $d_{model} \times 4d_{model}$ layer with 24 blocks and a $d_{model}$ value of 1024. The second and third rows contrast the results of the three-layer networks in Table \ref{results}, demonstrating that the base configuration is not overfitting the training data, though the result is not statistically significant at the 5\% level for the shallower network on the Booksum data set.

The fourth and fifth rows reduce the FFN multiple from $4d$ to $2d$, resulting in linear layers of $2d_{model} \times d_{model}$ and $d_{model} \times 2d_{model}$. The fifth row also increases the value of $d_{model}$ to roughly have the same parameter count as the first row. The standard FFN size outperforms these configurations on both datasets, though the difference is not statistically significant at the 5\% level for the $d_{model}=1248$ configuration on the Wikitext data set. 

\section{Conclusions}
Our results confirm that linear layers are important for the performance of transformer-based models. Based on our findings, shallower networks with a third layer within the transformer block FFNs yield improved cross-entropy loss while taking less time to train compared to the standard two-layer transformer block configuration. 

From here, additional work examining even larger FFNs could be valuable. Our results show that models with more linear layers outperform equivalent models with fewer layers. Future work could focus on larger networks, either by adding more linear layers or increasing the dimensionality of the FFN.

\bibliographystyle{plain}
\bibliography{references}

\end{document}